\documentclass[10pt,a4paper]{article}
\usepackage[T1]{fontenc}
\usepackage[utf8]{inputenc}
\usepackage{amssymb}
\usepackage[standard-baselineskips]{cmbright}
\usepackage[sfdefault,semibold,type1]{plex-sans}
\usepackage[tracking=true]{microtype}
\SetTracking{encoding={OT1,OML,OMS,OMX,U}}{0}
\SetTracking{encoding=T1,family={IBMPlexSans-TLF,plxSans-TLF}}{-15}
\SetTracking{encoding=TS1}{0}
\DeclareMathSizes{10.3}{11.124}{7.7868}{5.562}
\DeclareMathSizes{7}{7.56}{5.4}{5.4}
\DeclareMathSizes{8}{8.64}{6.48}{5.4}
\DeclareMathSizes{9}{9.72}{6.48}{5.4}
\DeclareMathSizes{10}{10.8}{7.56}{5.4}
\DeclareMathSizes{12}{12.96}{8.64}{6.48}
\AtBeginDocument{\lsstyle}
\usepackage[a4paper,top=17mm,bottom=19mm,left=21mm,right=21mm,
            headheight=19pt,headsep=5mm,footskip=10mm]{geometry}

\usepackage{graphicx}
\usepackage{amsmath}
\usepackage{siunitx}
\usepackage{booktabs,tabularx,array,multirow}
\usepackage{threeparttable}
\usepackage{makecell}
\usepackage{subcaption}
\usepackage{placeins}
\usepackage{float}
\usepackage{enumitem}
\usepackage{titlesec}
\usepackage{fancyhdr}
\usepackage[most]{tcolorbox}
\usepackage{needspace}
\usepackage[numbers,sort&compress]{natbib}
\providecommand{\bibcommenthead}{}
\usepackage{xcolor}
\usepackage{xurl}
\usepackage{hyperref}

\definecolor{SignalInk}{HTML}{24313A}
\definecolor{SignalHeading}{HTML}{1C236C}
\definecolor{SignalBar}{HTML}{1C236C}
\definecolor{SignalTeal}{HTML}{335DD1}
\definecolor{SignalDarkTeal}{HTML}{1C236C}
\definecolor{SignalMist}{HTML}{DCEAFE}
\definecolor{SignalLine}{HTML}{DCEAFE}
\definecolor{SignalMuted}{HTML}{60717B}

\hypersetup{colorlinks=true,linkcolor=SignalTeal,citecolor=SignalDarkTeal,
  urlcolor=SignalDarkTeal,pdfdisplaydoctitle=true}

\color{SignalInk}
\renewcommand{\arraystretch}{1.18}

\setlist{leftmargin=*,topsep=3pt,itemsep=2pt}
\let\PreprintBeginFigure\figure
\let\PreprintEndFigure\endfigure
\renewenvironment{figure}[1][!htbp]
  {\PreprintBeginFigure[htb]}{\PreprintEndFigure}


\renewcommand{\headrulewidth}{0.35pt}
\renewcommand{\headrule}{\hbox to\headwidth{\color{SignalLine}\leaders\hrule height \headrulewidth\hfill}}

\titleformat{\section}
  {\needspace{4\baselineskip}\Large\bfseries\color{SignalHeading}}
  {\color{SignalTeal}\makebox[9mm][l]{\thesection}}
  {0pt}{}
  [\vspace{3pt}{\color{SignalLine}\titlerule}]
\titleformat{name=\section,numberless}
  {\needspace{4\baselineskip}\Large\bfseries\color{SignalHeading}}
  {}{0pt}{}
  [\vspace{3pt}{\color{SignalLine}\titlerule}]
\titlespacing*{\section}{0pt}{2.0em}{0.75em}

\titleformat{\subsection}
  {\needspace{3\baselineskip}\large\bfseries\color{SignalHeading}}
  {\color{SignalTeal}\thesubsection}
  {0.65em}{}
\titleformat{name=\subsection,numberless}
  {\needspace{3\baselineskip}\large\bfseries\color{SignalHeading}}
  {}{0pt}{}
\titlespacing*{\subsection}{0pt}{1.45em}{0.35em}

\titleformat{\subsubsection}
  {\needspace{2.5\baselineskip}\normalsize\bfseries\color{SignalHeading}}
  {\color{SignalTeal}\thesubsubsection}
  {0.65em}{}
\titleformat{name=\subsubsection,numberless}
  {\needspace{2.5\baselineskip}\normalsize\bfseries\color{SignalHeading}}
  {}{0pt}{}
\titlespacing*{\subsubsection}{0pt}{1.2em}{0.25em}

\let\PaperOriginalTable\table
\let\PaperOriginalEndTable\endtable
\renewenvironment{table}[1][tbp]
  {\PaperOriginalTable[htb]}
  {\PaperOriginalEndTable}

\usepackage{xparse}
\usepackage{mathrsfs,textcomp,algorithm,algorithmicx,algpseudocode,listings}
\newcommand{\PaperAuthorsDisplay}{}
\newcommand{\PaperAffiliationsDisplay}{}
\newcommand{\PaperCorrespondence}{}

\newcommand{\PaperTitle}{}
\newcommand{\fnm}[1]{#1}
\newcommand{\sur}[1]{#1}
\newcommand{\orgname}[1]{#1}
\newcommand{\orgdiv}[1]{#1}
\newcommand{\orgaddress}[1]{#1}
\newcommand{\city}[1]{#1}
\newcommand{\country}[1]{#1}
\RenewDocumentCommand{\title}{o m}{\gdef\PaperTitle{#2}}

\NewDocumentCommand{\affil}{O{} m}{%
  \gappto\PaperAffiliationsDisplay{\textsuperscript{#1}#2\par}}
\NewDocumentCommand{\email}{m}{\gappto\PaperCorrespondence{#1\par}}

\AddToHook{begindocument/before}{\definecolor{SignalInk}{HTML}{2F3135}
\definecolor{SignalMuted}{HTML}{5B637A}
\definecolor{SignalHeading}{HTML}{1C236C}
\definecolor{SignalBar}{HTML}{335DD1}
\definecolor{SignalMist}{HTML}{F7F9FF}
\definecolor{PaperBackground}{HTML}{FCFDFF}
\pagecolor{PaperBackground}
\color{SignalInk}
\urlstyle{same}
\setlength{\parskip}{0.42em plus 0.08em minus 0.05em}
\clubpenalty=10000
\widowpenalty=10000
\displaywidowpenalty=10000
\setlist[enumerate]{beginpenalty=10000,midpenalty=10000,endpenalty=0}

\AddToHook{cmd/section/before}{\Needspace{6\baselineskip}}

\titleformat{\section}
  {\Large\bfseries\color{SignalHeading}}
  {\makebox[9mm][l]{\thesection}}{0pt}{}
\titleformat{name=\section,numberless}
  {\Large\bfseries\color{SignalHeading}}
  {}{0pt}{}
\titleformat{\subsection}
  {\large\bfseries\color{SignalHeading}}
  {\thesubsection}{0.65em}{}
\titleformat{\subsubsection}
  {\normalsize\bfseries\color{SignalHeading}}
  {\thesubsubsection}{0.65em}{}

\captionsetup{
  font=small,
  labelfont={bf,color=SignalTeal},
  textfont={color=SignalMuted},
  labelsep=period,
  justification=raggedright,
  singlelinecheck=false,
  skip=8pt
}
\fancyhead[L]{\footnotesize\color{SignalMuted}U-NeXt: neonatal EEG segmentation}
\fancyhead[R]{\footnotesize\color{SignalMuted}Research article}

\newif\ifPaperStyleFirstAuthor
\PaperStyleFirstAuthortrue
\newcount\PaperStyleAuthorCount
\RenewDocumentCommand{\author}{s O{} m}{%
  \global\advance\PaperStyleAuthorCount by 1
  \ifPaperStyleFirstAuthor\global\PaperStyleFirstAuthorfalse
  \else
    \ifnum\PaperStyleAuthorCount=5\gappto\PaperAuthorsDisplay{,\\}%
    \else\gappto\PaperAuthorsDisplay{,\space}\fi
  \fi
  \IfBooleanTF{#1}
    {\gappto\PaperAuthorsDisplay{#3\textsuperscript{#2*}}}
    {\gappto\PaperAuthorsDisplay{#3\textsuperscript{#2}}}}

\newcommand{\PaperStyleTag}[1]{%
  \tcbox[on line,colback=white!9!SignalBar,colframe=white!15!SignalBar,
    boxrule=0.3pt,arc=1mm,boxsep=0pt,left=2mm,right=2mm,top=1.4mm,bottom=1.4mm]{%
    \fontsize{7}{9}\selectfont\mdseries\color{white}\strut{}#1}%
}

\newcommand{\PaperStyleAbstractRow}[2]{%
  {\fontsize{7.5}{9.5}\selectfont\bfseries\color{SignalTeal}\MakeUppercase{#1}} &
  \fontsize{9}{11.5}\selectfont\color{SignalInk}#2\tabularnewline[5pt]
}

\renewcommand{\maketitle}{%
  \hypersetup{pdftitle={\PaperTitle},pdfauthor={Robert Hogan et al.},
    pdfsubject={U-NeXt: general-purpose event detection for neonatal EEG}}%
  \thispagestyle{empty}%
  \begingroup
  \setlength{\parskip}{0pt}%
  \vspace*{-10mm}%
  \begin{tcolorbox}[
    enhanced,width=\dimexpr\textwidth+6mm\relax,
    enlarge left by=-3mm,enlarge right by=-3mm,
    colback=SignalBar,colframe=SignalBar,boxrule=0pt,arc=1.5mm,
    boxsep=0pt,left=8mm,right=8mm,top=8mm,bottom=8mm,
    before skip=0pt,after skip=8mm]
    \color{white}%
    {\fontsize{7.5}{10}\selectfont
      \PaperStyleTag{Preprint\enspace\textbullet\enspace October 2026}\par}
    \vspace{8mm}%
    {\raggedright\hyphenpenalty=10000\fontsize{26}{29}\selectfont
      \fontseries{medium}\selectfont\color{white}\PaperTitle\par}
    \vspace{7mm}%
    {\raggedright\fontsize{10.3}{13.4}\selectfont\mdseries
      \color{white}\PaperAuthorsDisplay\par}
    \vspace{5mm}%
    \begin{tabularx}{\linewidth}{@{}X@{\hspace{6mm}}>{\raggedright\arraybackslash}p{46mm}@{}}
      \fontsize{7.5}{10}\selectfont\setlength{\parskip}{0pt}\lineskiplimit=-10pt\relax\color{white!90!SignalBar}\PaperAffiliationsDisplay &
      \fontsize{7.5}{10}\selectfont\setlength{\parskip}{0pt}\color{white!90!SignalBar}
      \textbf{Correspondence}\par\PaperCorrespondence
    \end{tabularx}
  \end{tcolorbox}
  \begin{tcolorbox}[
    colback=SignalMist,colframe=SignalMist,boxrule=0pt,arc=0pt,
    boxsep=0pt,left=5mm,right=5mm,top=4mm,bottom=2.5mm,
    before skip=0pt,after skip=0pt]
    {\fontsize{8}{10}\selectfont\bfseries\color{SignalHeading}ABSTRACT\par}
    \vspace{2mm}{\color{SignalLine}\hrule height 0.35pt}\vspace{3mm}
    \fontsize{9}{11.5}\selectfont
    \renewcommand{\arraystretch}{1}%
    \begin{tabularx}{\linewidth}{@{}p{21mm}@{\hspace{3mm}}X@{}}
      \PaperStyleAbstractRow{Background}{Automated neonatal EEG analysis can support continuous monitoring where specialist expertise is limited. Existing event detectors typically classify short EEG segments, restricting temporal context and limiting detection resolution. Methods are also commonly developed separately for individual event types.}
\PaperStyleAbstractRow{Objectives}{To develop a general framework that treats neonatal EEG event detection as a semantic segmentation task, improving temporal localisation, detection accuracy, and computational efficiency across seizures, low-voltage intervals (LVIs), and artefacts.}
\PaperStyleAbstractRow{Methods}{We developed U-NeXt, a convolutional encoder--decoder network that directly segments individual EEG channels, using longer EEG segments to preserve temporal context. A shared architectural design and training framework were developed using LVI, artefact, and seizure datasets from 51, 98, and 278 neonates, respectively, comprising over 11,000 hours of EEG and 82,000 annotated events. Performance was compared with task-specific ConvNeXt classifiers. For seizure detection, we also evaluated a model trained on an expanded dataset, U-NeXt+. Both U-NeXt variants and the ConvNeXt baseline were evaluated for seizure detection on two independent held-out datasets from Cork and Helsinki, comprising 130 neonates, with recordings annotated by three experts.}
\PaperStyleAbstractRow{Results}{U-NeXt improved detection across all three event types, with particular gains
in sensitivity for short events. The U-NeXt+ variant achieved Matthews correlation coefficients of
0.862 and 0.891 on Helsinki and Cork, respectively, and improved hourly seizure burden correlations by 30--40\% relative to ConvNeXt. U-NeXt+ met the criterion for expert-level performance on both datasets ($p$$<$0.001). In separate computational benchmarks, U-NeXt reduced batch inference time relative to ConvNeXt by 254-fold on CPU and 200-fold on GPU.}
\PaperStyleAbstractRow{Conclusion}{A common segmentation framework improves detection across diverse neonatal EEG event types while substantially reducing inference time. Accurate seizure burden estimates and performance comparable to expert reviewers support its potential use in continuous EEG monitoring and clinical decision support.}

    \end{tabularx}
  \end{tcolorbox}
  \vfill
  {\color{SignalTeal}\rule{\textwidth}{0.7pt}}\par
  \endgroup
  \clearpage
  \fontsize{10.3}{13.4}\selectfont
}
}

\NewHook{paper/layout/before-postprocessing}
\NewHook{paper/layout/before-training}
\NewHook{paper/layout/before-inference}
\NewHook{paper/layout/before-reduced-montage}

\newcommand{\PaperFigure}[1]{%
  \ifcsname PaperFigurePlaced@#1\endcsname\else
    \csname PaperFigureContent@#1\endcsname
    \expandafter\gdef\csname PaperFigurePlaced@#1\endcsname{}%
  \fi}

\expandafter\newcommand\csname PaperFigureContent@architecture\endcsname{%
\begin{figure}[!htbp]
    \centering
    \includegraphics[width=1.0\linewidth]{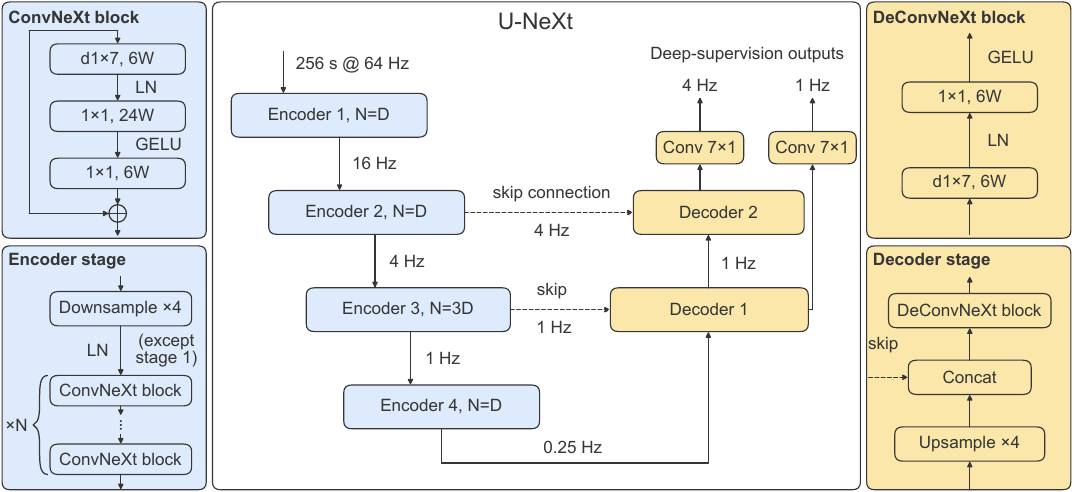}
    \caption{U-NeXt architecture. The 4-stage encoder (pale blue), adapted from ConvNeXt \cite{hogan2025scaling}, contains downsampling, normalisation, and convolution blocks. The 2-stage decoder (pale orange) contains upsampling and convolution blocks. The blocks use depthwise convolution ($d$), layer normalisation (LN), and Gaussian error linear unit (GELU) activation. Downsampling uses a $1\times4$ convolution; upsampling uses a $1\times4$ transposed convolution, both with stride 4. Skip connections concatenate encoder and decoder features and 2 decoder outputs provide deep supervision at different resolutions. $W$ and $D$ control width and depth, respectively.}
    \label{fig:arch}
\end{figure}
}

\expandafter\newcommand\csname PaperFigureContent@validation\endcsname{%
\begin{figure}
    \centering
    \includegraphics[width=1\linewidth]{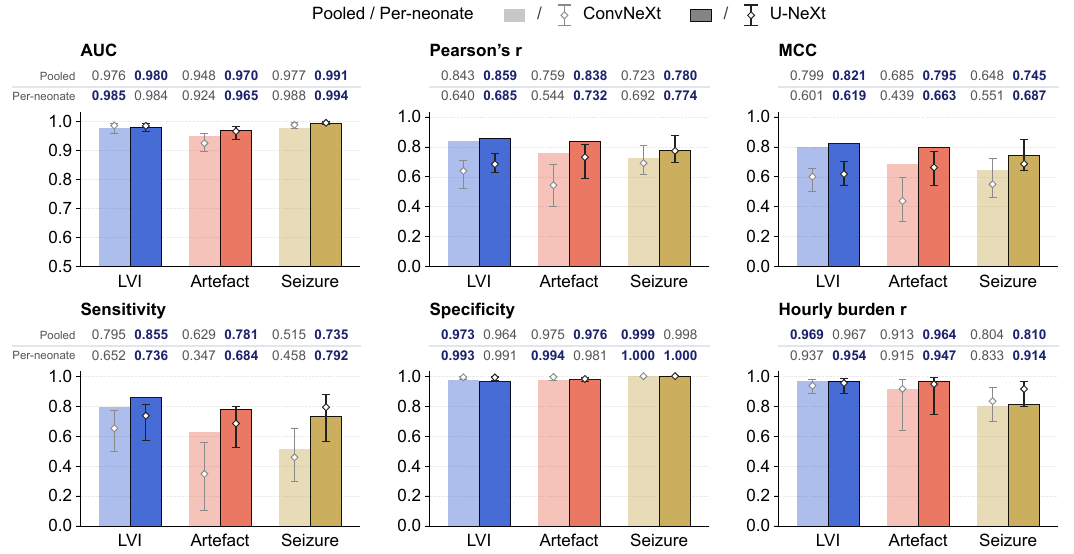}
    \caption{Comparison of U-NeXt and ConvNeXt on development-set validation data. Bars show pooled metrics; diamonds and whiskers show per-neonate medians and interquartile ranges. LVI used a Nano model with 10-fold cross-validation; artefact, Medium with 5-fold cross-validation; seizure, XL with a fixed 80:20 train/validation split. U-NeXt outperformed task-customised ConvNeXt baselines on every task despite using a common configuration.}
    \label{fig:cnx_vs_unet_testing}
\end{figure}
}

\expandafter\newcommand\csname PaperFigureContent@event-duration\endcsname{%
\begin{figure}
    \centering
    \includegraphics[width=1\linewidth]{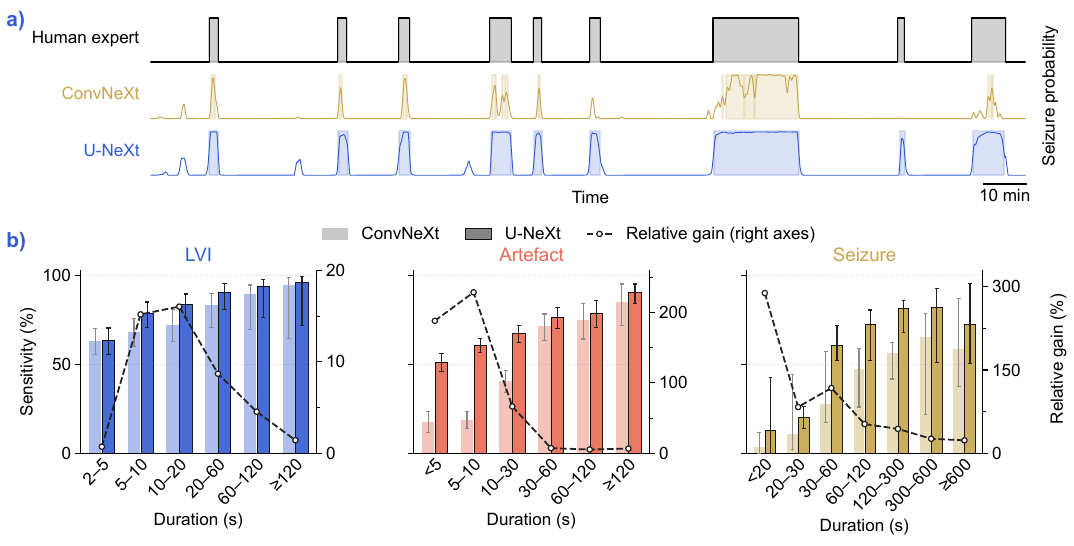}
    \caption{\panelref{a} ConvNeXt and U-NeXt seizure probabilities with expert annotation over 3.5 hours from one recording. Shading marks positive annotations or predictions at a 0.5 threshold; scale bar indicates 10 minutes. U-NeXt produces a more coherent trace, accurate event boundaries, and better short-event sensitivity. \panelref{b} Sample sensitivity by annotated event duration across the three task-specific datasets. Dashed lines show relative sensitivity gains on task-specific right axes. Error bars: 95\% percentile intervals from N=3,000 per-neonate bootstrap resamples.}
    \label{fig:seizure_duration}
    \label{fig:sz_probs}
\end{figure}
}

\expandafter\newcommand\csname PaperFigureContent@false-detections\endcsname{%
\begin{figure}[!ht]
    \centering
    \includegraphics[width=\linewidth]{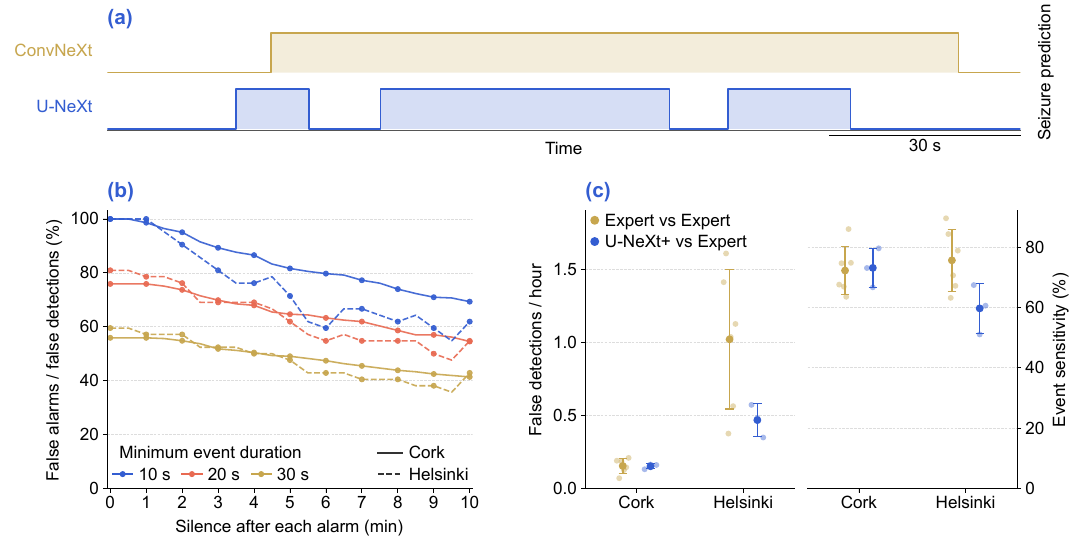}
    \caption{\panelref{a} False-positive fragmentation in a seizure-free interval. ConvNeXt produces one detection; U-NeXt produces three but with 29.6\% less false-positive time. The U-NeXt prediction is better but triples FD/h. \panelref{b} Impact of alarm configuration (minimum event duration and post-alarm silence period) on false alarms vs false detections in the prediction mask. Reasonable values can reduce false alarms by approximately 50\% relative to FD/h. \panelref{c} Comparison of inter-rater FD/h and event sensitivity with model performance when using the same expert reference. The points and error bars show the mean and standard deviation across reference-rater choices; individual data points are shown for each reference choice.}
    \label{fig:fp_fragmentation}
    \label{fig:false-alarms}
\end{figure}
}

\expandafter\newcommand\csname PaperFigureContent@inference\endcsname{%
\begin{figure}
    \centering
    \includegraphics[width=\linewidth]{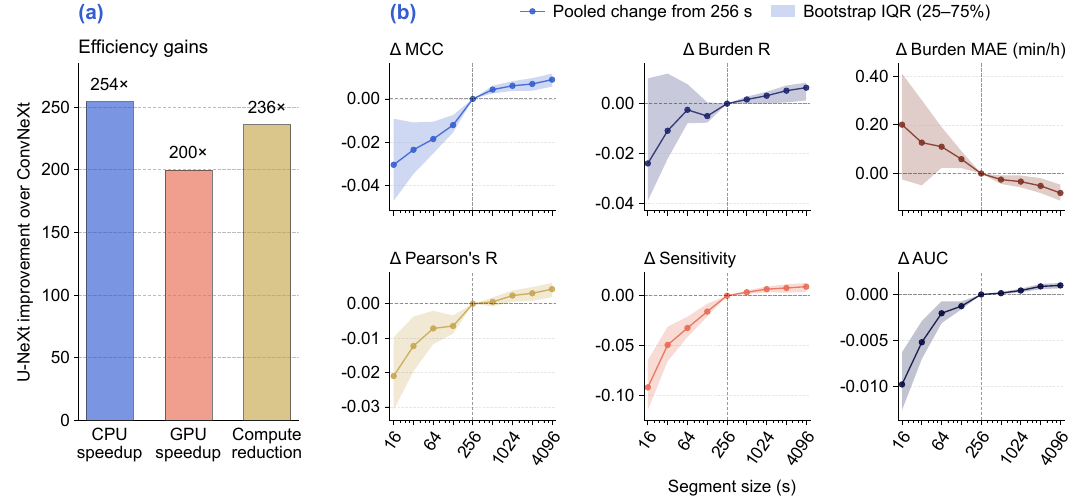}
    \caption{\small \panelref{a} U-NeXt inference-time and compute gains over ConvNeXt for seizure detection on 1 hour of 18-channel EEG from Helsinki. \panelref{b} Effect of input segment size on U-NeXt seizure detection on Helsinki, relative to the 256~s training length. Burden metrics are calculated hourly. Shading shows interquartile ranges from N=5,000 neonate bootstrap resamples, each relative to its own 256~s baseline. On the logarithmic segment-size axis, performance improves rapidly from very short segments with limited EEG context, then stabilises with modest gains for longer segments.}
    \label{fig:segment_size}
\end{figure}
}

\expandafter\newcommand\csname PaperFigureContent@reduced-montage\endcsname{%
\begin{figure}[!ht]
    \centering
    \includegraphics[width=\linewidth]{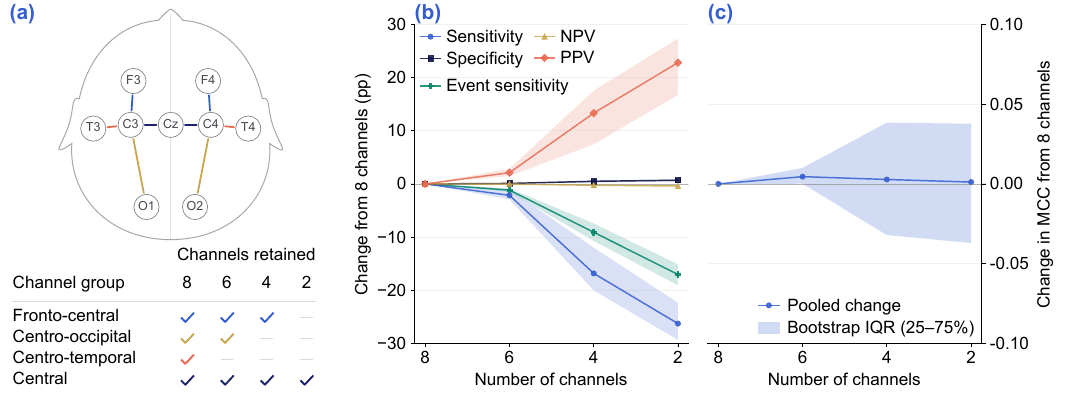}
    \caption{\small Reduced-montage performance of U-NeXt+ on Cork (51 neonates). \panelref{a} Bipolar derivations and channel groups retained in each montage. \panelref{b} Changes in sample-based sensitivity, specificity, NPV, PPV, and event sensitivity, in percentage points. With two central channels, event sensitivity is 82.2\% of its eight-channel value. A reference event is detected if it overlaps any predicted positive interval on the unanimous-consensus timeline. \panelref{c} Change in MCC. Changes are relative to the eight-channel reference montage. All configurations use maximum aggregation across retained channels. Shading shows pointwise interquartile range from $N=5,000$ neonate bootstrap resamples, each compared with its own eight-channel result.}
    \label{fig:reduced-montage}
\end{figure}
}

\AddToHook{paper/layout/before-training}{\PaperFigure{validation}}
\AddToHook{paper/layout/before-inference}{\PaperFigure{inference}}
\newcommand{\PreprintTopFigure}[1]{%
  \begingroup
  \renewenvironment{figure}[1][]{\PreprintBeginFigure[t]}{\PreprintEndFigure}%
  \PaperFigure{#1}%
  \endgroup}
\AddToHook{paper/layout/before-reduced-montage}{\PreprintTopFigure{reduced-montage}}

\newcommand{\PaperFundingStatement}{This work was supported under Grant DT~2023~435 by the Disruptive Technologies Innovation Fund (DTIF), administered by Enterprise Ireland,}
\newcommand{\PaperFunderRole}{which had no role in study design, data collection, analysis or interpretation, or manuscript preparation.}
\newcommand{\PaperCergenXDisclosure}{All authors are affiliated with CergenX Ltd., a University College Cork spin-out developing neonatal neuromonitoring technologies: G.B.\ and S.G.\ are co-founders; R.H., A.L.\ and J.O.T.\ are employees; and S.M.\ and S.V.\ were independent consultants.}

\newcommand{\PaperOtherInterests}{G.B.\ and S.G.\ are founders of Kephala Ltd., and S.M.\ is a consultant. G.B.\ also provides consultancy to Realta Lifesciences through University College Cork.}
\newcommand{\PaperCompetingInterests}{\PaperCergenXDisclosure\space\PaperOtherInterests}

\begin{document}
\fontsize{10.3}{13.4}\selectfont

\title[U-NeXt: General-purpose event detection for neonatal EEG]{U-NeXt: General-purpose event detection for neonatal EEG}

\author*[1]{\fnm{Robert} \sur{Hogan}}\email{rhogan@cergenx.com}

\author[1,2]{\fnm{Sean R.} \sur{Mathieson}}
\author[1]{\fnm{Aurel} \sur{Luca}}
\author[1,2]{\fnm{Soraia} \sur{Ventura}}
\author[1]{\fnm{Sean} \sur{Griffin}}
\author[1,2,3]{\fnm{Geraldine B.} \sur{Boylan}}
\author*[1]{\fnm{John M.} \sur{O'Toole}}\email{jotoole@cergenx.com}

\affil[1]{\orgname{CergenX Ltd}, \orgaddress{\city{Cork}, \country{Ireland}}}
\affil[2]{\orgdiv{INFANT Research Centre}, \orgname{University College Cork}, \orgaddress{\country{Ireland}}}
\affil[3]{\orgdiv{Department of Paediatrics and Child Health}, \orgname{University College Cork}, \orgaddress{\country{Ireland}}}

\maketitle

\definecolor{PanelLabelBlue}{HTML}{335DD1}
\DeclareRobustCommand{\panelref}[1]{\textcolor{PanelLabelBlue}{\textsf{\textbf{(#1)}}}}
\hypersetup{colorlinks=true,linkcolor=PanelLabelBlue}
\DeclareRobustCommand{\figref}[2][]{\hyperref[#2]{\textcolor{PanelLabelBlue}{Figure~\ref*{#2}#1}}}
\DeclareRobustCommand{\tabref}[1]{\hyperref[#1]{\textcolor{PanelLabelBlue}{Table~\ref*{#1}}}}

\section{Introduction}\label{sec:Intro}
Neonatal brain injury in the hours and days after birth remains a serious concern for the estimated 8.5 per 1000 infants born globally with  neonatal encephalopathy (NE)\cite{lee2013intrapartum}. The negative impacts of NE on neurodevelopmental outcome make it the
second leading neurological disorder contributing to disability-adjusted life years (DALYs)
\cite{steinmetz2024global}. Early diagnosis and treatment are paramount to improving lifelong neurodevelopmental outcome.

Continuous monitoring with electroencephalography (EEG) is the gold standard for the assessment of neonatal brain function. It is a relatively inexpensive and portable technology, but interpretation difficulty has limited widespread adoption. Neonatal EEG review is a highly specialised skill that is absent in many clinical settings, particularly in low- and middle-income countries where the burden of NE is particularly high \cite{kukka2022incidence}. In recent years there has been increasing interest in automated systems to reduce laborious expert EEG review in the NICU \cite{proietti2024advances,chioma2023applications,keles2023past,sullivan2024transforming}. In particular, we have recently shown that scaling deep learning methods can lead to expert-level retrospective analysis \cite{hogan2025scaling,otoole2026expert}.

\PaperFigure{architecture}

To date, most research in this area has focused on models for a single event detection task, often trained and evaluated on a single dataset \cite{Raurale2020,temko2011performance,webb2021automated}. This leads to highly specialised models that are tuned for that particular task--dataset combination and often use a specific electrode array or montage \cite{gramacki2022framework}. Datasets are often small, with performance evaluated through cross-validation, seldom including independent held-out validation \cite{Raurale2020,temko2011performance}. Insufficient evaluation means generalisation performance is unclear at best. At worst, methods overfit to particular task--dataset pairings, obscuring potential downstream utility and hindering clinical translation. In addition, each neonatal EEG event type requires its own study, making it difficult to build robust and generalisable methods.

EEG event detection commonly uses short-time classification, assigning one label per segment (e.g. 16~s for seizures). Post-processing then assembles these detections into a probability trace and binary mask. This approach has several limitations. First, the time resolution of the prediction mask is dependent on the length of the segments. Overlap percentage controls the sample rate, but not the resolution, although a high overlap is needed to capture the start and end times. A longer segment may make the event easier to detect but harder to localise in time \cite{Seeuws2024}. Selecting a segment size is therefore a compromise for events with widely varying durations, e.g. artefacts \cite{stevenson2014artefact,webb2021automated}. Second, short-time classification lacks the continuous temporal context experts use to annotate event boundaries. Treating variable-duration event detection as fixed-segment classification rather than segmentation therefore limits both context and temporal localisation.

In this work, we address these limitations by introducing a new general framework for event detection in neonatal EEG. We develop a new model based on a U-Net style architecture which directly learns the semantic segmentation task for 3 distinct applications: detection of seizures, low-voltage intervals (LVIs), and artefacts. Importantly, we do not customise the architecture or training recipe for each case and the design is based on single-channel input. This ensures that the framework is generalisable to other event types, independent of EEG montage. We trained and validated a model for each task using the development datasets, then evaluated seizure detection on the held-out datasets Cork and Helsinki. Training a high-quality detector for this diverse set of tasks with a single unified architecture is a significant challenge but an important step towards more general and scalable approaches to AI for neonatal EEG. The history of AI development has repeatedly shown that detailed customisation of models to each task may yield short-term gains but will ultimately be shown to be more brittle than the general systems that inevitably surpass them.

\section{Methods}\label{sec:Methods}
We adapted our state-of-the-art 1D ConvNeXt seizure classifier \cite{hogan2025scaling} for segmentation using a U-Net encoder, bottleneck, and decoder with encoder--decoder skip connections.
The model width and depth can be adjusted with two parameters to suit the dataset size.

We apply the same model and optimiser hyperparameters to three event detection tasks, namely seizure, artefact, and LVI detection.
Event durations and signal morphologies in all three tasks vary considerably.
Seizures annotated in this study are characterised by repetitive EEG patterns that evolve over the event duration \cite{tsuchida2013terminology}.
They can be represented as harmonic signals with slowly varying instantaneous frequency, amplitude modulation, and morphology.
Event duration can range from 10~s to many minutes, with a median duration of approximately 2 minutes \cite{PavelAnSer}.
LVIs are events that are $\geq$ 2~s with peak-to-peak amplitude $<$25 $\mu$V.
Artefacts are the most heterogeneous group, reflecting their varied sources. They include regular patterns, such as periodic respiration artefact, and stochastic activity, such as muscle artefacts resembling band-limited noise.
Artefact durations range from $<$1~s to hours, complicating short-time classification.
Typical segment sizes for classification are from 8 to 32~s for the seizure task \cite{Stevenson2019,hogan2025scaling}, 2~s for the LVI task \cite{otoole2026expert}, and between 4 and 64~s \cite{stevenson2014artefact,webb2021automated} for artefact.
For all three tasks, 256~s input segments accommodate varied event durations and let the model learn from longer temporal context, similar to human review.

\subsection{U-NeXt architecture}
Our proposed U-NeXt architecture is a U-Net model based on the ConvNeXt backbone described in \cite{hogan2025scaling}. U-NeXt uses the ConvNeXt encoder without its head (\figref{fig:arch}) and with downsampling now $4\times$ between all stages, rather than $4\times$ after the stem and $2\times$ thereafter \cite{hogan2025scaling}. We then add a lightweight decoder stage for upsampling. Two integers control the architecture: $D$ sets the blocks per stage (depth), and $W$ sets each block's feature dimension (width). In this work, we use 3 variants (see \tabref{tab:model_variants}): Nano ($D=1, W=1$), Medium ($D=3, W=4$), and XL ($D=6, W=10$).

Each decoder stage comprises an upsampling layer, concatenation with an encoder skip connection, and a DeConvNeXt block based on a simplified ConvNeXt block (see \figref{fig:arch}). Note
also that the decoder scales with $W$ but does not scale with $D$. More complex decoders, such as mirrored ConvNeXt block structures or additional DeConvNeXt blocks, increased parameters and FLOPs without reliable performance gains.

Predicting annotation masks at the 64~Hz input resolution exceeds annotation precision and risks overfitting. To address this, we made 3 modifications:
\begin{enumerate}[leftmargin=2em,labelsep=0.5em]
    \item increased the downsampling between encoder stages,
    \item reduced the number of decoder stages,
    \item used deep supervision of the outputs of both decoder stages.
\end{enumerate}
These changes limited the maximum output resolution to 4~Hz, stabilised learning, and yielded good generalisation performance.

\begin{table}
    \centering
    \small
    \caption{\small Model size and compute for U-NeXt and equivalent ConvNeXt models \cite{hogan2025scaling}. Floating-point operation (FLOP) counts use one operation per multiply--add; U-NeXt uses 16\ s inputs for direct comparison.}
    \label{tab:model_variants}
      \begin{tabular}{lccrrrr}
        \toprule
                    &       &       &  \multicolumn{2}{c}{ConvNeXt} &   \multicolumn{2}{c}{U-NeXt}     \\
        \cmidrule(lr){4-5} \cmidrule(lr){6-7}
        model       & depth ($D$) & width ($W$) & \# params & FLOP  & \# params & FLOP \\
        \midrule
        Nano        & 1     & 1     & 38.7 k     & 1.9 m    & 50.1 k &  638.78 k \\
        Medium      & 3     & 4     & 1.7 m      & 84.3 m   & 1.9 m  &   23.8 m \\
        XL          & 6     & 10    & 20.6 m     & 1 G      & 21.7 m &  280.3 m  \\
        \bottomrule
    \end{tabular}
\end{table}

\subsection{Tasks and datasets}
We test this method on 3 tasks: detection of seizures, LVIs, and artefacts. These events cover a wide range of durations, amplitudes, frequency content, and characteristic morphologies, providing a broad assessment of neonatal EEG event detection. Good performance across these diverse tasks suggests that the framework may also be suitable for other event types.

For model development, we used EEG records from 329 term neonates in the fully anonymised NeoBase neonatal EEG database. EEG recordings in NeoBase were acquired during research studies at Cork University Maternity Hospital (CUMH), Ireland and University College London Hospital (UCLH), UK. During the original research studies, ethical approval was obtained from the relevant clinical research ethics committee and informed consent was obtained from the parents or guardians for the use of fully anonymised data in future research studies. In addition, ethical approval for secondary use of these datasets was obtained from the Clinical Research Ethics Committee of the Cork Teaching Hospitals in Ireland. EEG recordings began as soon as possible after birth and lasted hours or days using several different EEG acquisition machines (see \cite{hogan2025scaling} for details). EEGs were recorded from neonates with mixed aetiologies who were at risk of seizures in the neonatal intensive care unit (NICU) as well as a smaller group of healthy term newborns recorded in the postnatal wards ($\leq 2$~hours in recording duration) \cite{Korotchikova2009,Korotchikova2016,Raurale2020,Garvey2021}.

This EEG data was used to create 3 task-specific development datasets (see \tabref{tab:clin_demographics}). Two of the datasets used, seizure and LVI, have been previously described in \cite{hogan2025scaling} and \cite{otoole2026expert}, respectively; the artefact dataset was created for this work. Multi-channel EEG was reviewed by a neonatal neurophysiologist\footnote{Authors SRM and SV reviewed the seizure dataset. Author SRM reviewed the LVI dataset. Author SV reviewed the artefact dataset.} in each case, and events were annotated on a per-channel basis. For seizures, the entire recording was reviewed. For LVI and artefact, epochs of 1 and 3 hours, respectively, were reviewed. In the case of artefact, only the 2 central channels (C4--CZ, CZ--C3) were annotated with binary labels (artefact/no artefact).

Compared to \cite{hogan2025scaling}, we have expanded the seizure development data with long-duration, fully anonymised EEGs from a further 76 neonates recorded at UCLH, annotated per channel by a single neurophysiologist. We refer to this additional cohort as "Seizure+" (see \tabref{tab:clin_demographics}). Combining the original Seizure dataset ($n$=202) with Seizure+ gives 278 neonates, 10,893 hours of multi-channel EEG, and 24,432 per-channel seizure events. U-NeXt+ was trained on this combined dataset. For direct comparison with our previous ConvNeXt model, we also trained a U-NeXt model using the original 202-neonate dataset.

For seizure evaluation, we used EEG recordings from 130 neonates in the same two held-out datasets as in \cite{hogan2025scaling}, referred to throughout as Helsinki and Cork. Helsinki is the publicly available dataset ($n$=79; 112 hours; 18 channels) \cite{Stevenson2019}. Cork is a subset of NeoBase ($n$=51; 1,572 hours; 8 channels) recorded at CUMH. Both datasets have global seizure annotations from 3 independent reviewers. Neither the EEG data nor the reviewers overlap with those used for seizure development.

\begin{table}
  \centering
    \footnotesize
    \newcommand{\demolinegap}{-2pt}
    \ifdim\linewidth<400pt
      \let\scriptsize\footnotesize
      \renewcommand{\demolinegap}{0pt}
    \fi
    \caption{\small Cohort demographics: median (interquartile range) or number (percentage), unless stated otherwise; number of neonates ($n$) indicates incomplete data. LVI: low-voltage interval; HIE: hypoxic-ischaemic encephalopathy, with severity counts ordered mild/moderate/severe/unknown. The normal cohort comprises healthy postnatal-ward neonates \cite{Korotchikova2016}. EEG duration is per neonate; start is postnatal age at the first included EEG segment.}
  \label{tab:clin_demographics}
  \setlength{\tabcolsep}{0pt}
  \renewcommand{\arraystretch}{1.05}
  \newcommand{\demorange}[2]{\makecell[t]{#1\\[\demolinegap]{\scriptsize #2}}}
  \begin{tabular}{@{}p{.28\linewidth}*{4}{>{\centering\arraybackslash}p{.18\linewidth}}@{}}
    \toprule
     & \makecell{\textbf{LVI}\\($n=51$)} & \makecell{\textbf{Artefact}\\($n=98$)} & \makecell{\textbf{Seizure}\\($n=202$)} & \makecell{\textbf{Seizure+}\\($n=76$)} \\
    \midrule
    \mbox{Gestational age, weeks+days} & \demorange{40+2}{(39+2 to 40+5)} & \makecell[t]{39+6\\[\demolinegap]{\scriptsize (38+4 to 40+4)}\\[\demolinegap]{\scriptsize\color{gray} $n=97$}} & \makecell[t]{40+0\\[\demolinegap]{\scriptsize (39+2 to 40+5)}\\[\demolinegap]{\scriptsize\color{gray} $n=197$}} & \demorange{40+0}{(39+1 to 41+0)} \\
    \addlinespace[1pt]
    Birth weight, kg & \makecell[t]{3.50\\[\demolinegap]{\scriptsize (3.33--3.85)}\\[\demolinegap]{\scriptsize\color{gray} $n=26$}} & \makecell[t]{3.31\\[\demolinegap]{\scriptsize (2.32--3.57)}\\[\demolinegap]{\scriptsize\color{gray} $n=67$}} & \makecell[t]{3.50\\[\demolinegap]{\scriptsize (3.27--3.76)}\\[\demolinegap]{\scriptsize\color{gray} $n=77$}} & -- \\
    \addlinespace[1pt]
    Sex, female & 18 (35.3\%) & \makecell[t]{44 (45.4\%)\\[\demolinegap]{\scriptsize\color{gray} $n=97$}} & \makecell[t]{86 (42.8\%)\\[\demolinegap]{\scriptsize\color{gray} $n=201$}} & -- \\
    \addlinespace[0pt]
    \multicolumn{5}{@{}l}{\textbf{Clinical demographics}} \\
    \midrule
    Primary diagnosis & & & & \\
    \hspace{.7em}Normal cohort & 25 (49.0\%) & 0 (0\%) & 73 (41.0\%) & -- \\
    \hspace{.7em}HIE & 26 (51.0\%) & 43 (51.8\%) & 63 (35.4\%) & 45 (59.2\%) \\
    \hspace{1.4em}Severity & 9 / 13 / 4 / 0 & 18 / 22 / 3 / 0 & 23 / 26 / 14 / 0 & 15 / 17 / 8 / 5 \\
    \hspace{.7em}Birth asphyxia & -- & 5 (6.0\%) & 7 (3.9\%) & 7 (9.2\%) \\
    \hspace{.7em}Stroke & -- & 5 (6.0\%) & 9 (5.1\%) & 3 (3.9\%) \\
    \hspace{.7em}Other & -- & \makecell[t]{9 (10.8\%)\\[\demolinegap]{\scriptsize\color{gray} $n=83$}} & \makecell[t]{9 (5.1\%)\\[\demolinegap]{\scriptsize\color{gray} $n=178$}} & 21 (27.6\%) \\
    \addlinespace[1pt]
    Hypothermia & 18 (35.3\%) & \makecell[t]{17 (37.0\%)\\[\demolinegap]{\scriptsize\color{gray} $n=46$}} & \makecell[t]{28 (18.7\%)\\[\demolinegap]{\scriptsize\color{gray} $n=150$}} & -- \\
    \addlinespace[0pt]
    \multicolumn{5}{@{}l}{\textbf{EEG characteristics}} \\
    \midrule
    Total duration, h & 80.3 & 294 & 6,487 & 4,406 \\
    \addlinespace[1pt]
    Duration, h & \mbox{1.03 (1.00--2.00)} & \mbox{3.0 (3.0--3.0)} & \mbox{7.4 (1.1--51.6)} & \mbox{40.9 (20.2--88.3)} \\
    \addlinespace[1pt]
    Start, h & \mbox{23.0 (12.0--36.0)} & \mbox{37.5 (17.3--63.0)} & \mbox{13.8 (10.3--27.8)} & \mbox{10.0 (6.0--22.0)} \\
    \addlinespace[1pt]
    Neonates w/ seizures & 6 (11.8\%) & \makecell[t]{22 (26.5\%)\\[\demolinegap]{\scriptsize\color{gray} $n=83$}} & 77 (38.1\%) & 28 (36.8\%) \\
    Annotated events & 36,580 & 21,105 & 12,402 & 12,030 \\
    \bottomrule
  \end{tabular}
\end{table}

\subsection{Pre-processing and data augmentation}
Pre-processing of the EEG consisted of bandpass filtering within the 0.3--30 Hz passband and downsampling to 64~Hz. EEG was divided into 300~s segments with no overlap. During training, a random crop of 256~s was taken at each iteration. The training data was further dynamically augmented with flipping, magnitude jitter, and cutout \cite{devries2017cutout}. Augmentation hyperparameters were fixed across datasets.

We split each development dataset by neonate into training and validation sets. For the seizure dataset we used a single 80:20 split. For LVI and artefact we performed 10- and 5-fold cross-validation, respectively. For held-out seizure evaluation on Helsinki \cite{Stevenson2019} and Cork, we retrained each model on its full seizure development dataset.

The baseline ConvNeXt models followed similar pre-processing and data augmentation schemes but were
specialised to their respective datasets, as described in \cite{otoole2026expert, hogan2025scaling}. For example,
they used different data augmentations, hyperparameters, and kernel sizes. The ConvNeXt artefact baseline followed the same recipe as the ConvNeXt seizure model. ConvNeXt used 2~s, 8~s, and 16~s segments for LVI, artefact, and seizures, respectively.

\UseHook{paper/layout/before-postprocessing}
\subsection{Post-processing}\label{sec:post-processing}

In \cite{hogan2025scaling} we tried to use a minimal approach to post-processing, but recovering a high-resolution prediction from classification of 16~s segments required $>$98\% segment overlap and smoothing with a 32~s window. We argued that this step may improve performance when tuned to a particular benchmark dataset or metric, but will in most cases impair performance on another dataset or metric. With U-NeXt the direct output of a 4~Hz prediction mask (compared to 1/16~Hz in \cite{hogan2025scaling}) removes the need for overlap and reduces the importance of smoothing. Keeping a 16~s smoothing window provided limited benefit for seizure detection but we dropped it for the other models. We also removed the filling of short gaps between events. The only remaining post-processing step is to remove predictions shorter than the minimum event durations ($<10$~s for seizures; $<2$~s for LVIs; no filter for artefacts). The LVI ConvNeXt baseline retained 1~s rectangular smoothing, filling of gaps shorter than 1~s, and removal of events shorter than 1~s \cite{otoole2026expert}.

Without the need for inference overlap, each second of EEG passes through the model only once. Combined with a more efficient forward pass (\tabref{tab:model_variants}), this substantially speeds inference (Section \ref{sec:Results}).

\UseHook{paper/layout/before-training}
\subsection{Training}
Semantic segmentation uses diverse loss functions \cite{azad2023loss}, typically chosen for the data distribution (e.g. imbalance) and relative cost of errors (e.g. missing short features). To develop a general-purpose detector, we apply the same method across our 3 datasets despite their differing properties. LVIs are typically much shorter than seizures, with more frequent small pauses, while artefacts span both short and long timescales. We therefore construct a loss function that performs well in all of these scenarios using a combination of sample-wise binary cross entropy (BCE) and a symmetric Dice loss.

\begin{equation}
\text{Loss} = \sum_{o}{w_o \left(\alpha \text{BCE}(y, \hat{y})+ (1-\alpha)\text{Dice}_{sym}(y, \hat{y}) \right)}
\end{equation}
where $\text{Dice}_{sym}$ is a symmetric variant of the Dice loss given by
\begin{equation}
    \text{Dice}_{sym} = \frac{1}{2}\text{Dice}(y, \hat{y}) + \frac{1}{2} \text{Dice}(1-y, 1-\hat{y}),
\end{equation}
and
\begin{equation}
    \text{Dice}(y, \hat{y}) = 1 - \frac{2 y \odot \hat{y} +1}{y + \hat{y} +1},
\end{equation}
and $y, \hat{y}$ are the ground truth and model prediction respectively and $\odot$ denotes the Hadamard product. The sum runs over deep-supervision outputs $o$, with equal weights $w_o=0.5$ for the two outputs because alternative weighting schemes showed no clear advantage. The relative weight of the BCE and Dice is controlled by $\alpha$. The optimal value of this is likely dataset dependent, but for simplicity and to show the generality of the method, we fixed this to be 0.8 for all datasets.

The model was right sized for each task, using Nano for LVI, Medium for artefact, and XL for seizure detection (see \tabref{tab:model_variants}).

For optimisation we use AdamW with a single cycle of cosine annealing and 5\% warmup. We save an exponential moving average (EMA) of the model weights for improved performance and stability. Again, for generality, we fix most hyperparameters across all datasets using a batch size of 128 and $\beta =(0.9, 0.999)$. Weight decay is 0.05 for all tasks. The learning rate, EMA decay, and number of training epochs are adjusted to account for differences in datasets, model sizes, and convergence rates. We train for 500 epochs for LVI, 80 epochs for artefact, and 90 epochs for seizure. Finally, to mitigate the seizure dataset's approximately 50:1 imbalance, we sampled training batches with a maximum imbalance of 20:1; no other adjustments were made.

\section{Results}\label{sec:Results}

\subsection{Development set results}\label{sec:DevSetResults}
U-NeXt outperforms ConvNeXt on all three tasks (\figref{fig:cnx_vs_unet_testing}). For LVI, U-NeXt achieves a higher pooled MCC (0.821 vs 0.799), indicating better overall detection performance, although ConvNeXt performs better on some individual metrics. Notably, each ConvNeXt baseline was designed with task-specific input lengths, hyperparameters, augmentation, and even minor architectural adaptations such as kernel sizes. U-NeXt used almost the same configuration across datasets to maintain generalisability.

\PaperFigure{validation}

The superior performance across all event types and lack of customisation make it likely that the
U-NeXt framework will generalise well to other event types of interest. Furthermore, the performance
differences are most clear when there is larger headroom above the ConvNeXt models. As described in \cite{otoole2026expert} the ConvNeXt model already does exceptionally well at detecting LVIs so there is relatively little room for improvement. As the task complexity increases, however, the results demonstrate that U-NeXt is the more flexible and capable architecture in these regimes.

An example output illustrates the nature of the improvement over ConvNeXt (\figref[(a)]{fig:sz_probs}). ConvNeXt detects most seizures but produces a noisier trace and often misses event boundaries due to limited temporal context and output resolution. U-NeXt segments with broader EEG context, producing more coherent probabilities and accurate boundaries that better resemble human annotations. \figref[(b)]{fig:sz_probs} shows higher sample sensitivity across event durations, with larger relative gains for shorter events; performance is comparable for the shortest LVIs.

\PaperFigure{event-duration}

\subsection{Held-out seizure performance}
We next tested whether the gains observed during development translated to seizure detection on the Cork and Helsinki held-out datasets. \tabref{tab:seizure-performance} compares ConvNeXt and U-NeXt(+) on these two datasets. Since both datasets are only annotated globally, we need to aggregate the per-channel predictions. Although a simple maximum will result in the highest sensitivity we found that taking the mean of the top 2 channel probabilities before applying a standard $>0.5$ threshold gave better overall performance on both datasets. There are undoubtedly more sophisticated methods for learning a global pooling operation, which we do not explore here. For reference, we also included the open-source Helsinki SVM neonatal seizure detector \cite{tapani2019}\footnote{\raggedright Results were generated using the leave-one-baby-out models in the \href{https://github.com/ktapani/Neonatal_Seizure_Detection}{Helsinki neonatal seizure detector repository}. A variant is commercially available in the EU as part of the ANT Neo device.}. Unfortunately, it was impractical to run inference for this model on the larger Cork dataset (see Section \ref{sec:inference-performance} for further details).

\begin{table}[!p]
\centering
\caption{Seizure detection on held-out Helsinki and Cork datasets. U-NeXt models used whole Helsinki recordings and 10-minute Cork segments; Helsinki SVM used leave-one-neonate-out models. FP min/h: false-positive time per hour; FD/h: false detections per hour. Burden MAE: per-neonate mean absolute burden error (min/h). $\Delta\kappa$: change in inter-rater agreement when replacing one expert with the model; superscripts/subscripts give upper/lower 95\% bounds from 10,000 neonate bootstrap resamples. Bold marks best value with U-NeXt models significantly improving on ConvNeXt.}
\label{tab:seizure-performance}
\setlength{\tabcolsep}{0pt}
\renewcommand{\arraystretch}{1.02}
\begin{tabular*}{\linewidth}{@{\extracolsep{\fill}}lcccc@{}}
\toprule
 & Helsinki SVM & ConvNeXt & U-NeXt & U-NeXt+ \\
\midrule
\multicolumn{5}{@{}l}{\textbf{Helsinki (n=79)}} \\
\addlinespace[2pt]
MCC & 0.697 & 0.764 & \textbf{0.865} & 0.862 \\
Pearson's $r$ & 0.649 & 0.824 & \textbf{0.896} & 0.884 \\
AUC & 0.908 & 0.982 & \textbf{0.992} & 0.991 \\
Sens./Spec. (\%) & 73.2/96.7 & 72.9/98.5 & 83.7/\textbf{99.2} & \textbf{84.5}/99.0 \\
PPV/NPV (\%) & 72.7/96.8 & 84.9/96.8 & \textbf{92.2}/98.1 & 90.7/\textbf{98.2} \\
FP min/h & 1.606 & 0.758 & \textbf{0.417} & 0.516 \\
\noalign{\vskip 2pt\hrule height 0.25pt\vskip 2pt}
Event Sensitivity (\%) & 70.7 & 69.4 & 72.4 & \textbf{76.3} \\
FD/h & 1.036 & 0.459 & \textbf{0.225} & 0.274 \\
\noalign{\vskip 2pt\hrule height 0.25pt\vskip 2pt}
Burden $r$ (per neonate) & 0.790 & 0.810 & \textbf{0.968} & 0.957 \\
Burden $r$ (hourly) & 0.690 & 0.739 & 0.948 & \textbf{0.956} \\
Burden MAE: seizure ($n=39$) & 3.259 & 3.333 & \textbf{2.038} & 2.301 \\
Burden MAE: no seizure ($n=40$) & 2.214 & 1.370 & 0.623 & \textbf{0.542} \\
\noalign{\vskip 2pt\hrule height 0.25pt\vskip 2pt}
$\Delta\kappa$ (95\% CI) & $-0.109^{-0.028}_{-0.189}$ & $-0.082^{+0.002}_{-0.156}$ & $-0.022^{+0.022}_{-0.064}$ & $\mathbf{-0.021}^{+0.023}_{-0.061}$ \\
\midrule
\multicolumn{5}{@{}l}{\textbf{Cork (n=51)}} \\
\addlinespace[2pt]
MCC & \textemdash & 0.739 & 0.783 & \textbf{0.891} \\
Pearson's $r$ & \textemdash & 0.766 & 0.789 & \textbf{0.875} \\
AUC & \textemdash & 0.996 & 0.998 & \textbf{0.999} \\
Sens./Spec. (\%) & \textemdash & 88.9/99.2 & 92.8/99.3 & \textbf{94.7}/\textbf{99.7} \\
PPV/NPV (\%) & \textemdash & 62.0/99.8 & 66.6/\textbf{99.9} & \textbf{84.1}/\textbf{99.9} \\
FP min/h & \textemdash & 0.470 & 0.404 & \textbf{0.157} \\
\noalign{\vskip 2pt\hrule height 0.25pt\vskip 2pt}
Event Sensitivity (\%) & \textemdash & \textbf{94.9} & 93.5 & 93.6 \\
FD/h & \textemdash & 0.363 & 0.252 & \textbf{0.110} \\
\noalign{\vskip 2pt\hrule height 0.25pt\vskip 2pt}
Burden $r$ (per neonate) & \textemdash & 0.801 & 0.884 & \textbf{0.991} \\
Burden $r$ (hourly) & \textemdash & 0.667 & 0.754 & \textbf{0.937} \\
Burden MAE: seizure ($n=20$) & \textemdash & 0.714 & 0.491 & \textbf{0.159} \\
Burden MAE: no seizure ($n=31$) & \textemdash & 0.126 & 0.109 & \textbf{0.051} \\
\noalign{\vskip 2pt\hrule height 0.25pt\vskip 2pt}
$\Delta\kappa$ (95\% CI) & \textemdash & $-0.094^{+0.005}_{-0.189}$ & $-0.065^{+0.004}_{-0.130}$ & $\mathbf{-0.006}^{+0.012}_{-0.025}$ \\
\bottomrule
\end{tabular*}
\end{table}

U-NeXt improves substantially upon the previous state-of-the-art ConvNeXt model across a diverse set of metrics. Notably, we see large improvements in the estimation of seizure burden, with a 30--40\% relative improvement over ConvNeXt in the correlation between the predicted and true seizure burden. We also report seizure burden mean absolute error (MAE), the mean absolute difference between predicted and reference burden. For U-NeXt+ on Cork, this is $<$10~s/h for neonates with seizures and approximately 3~s/h for those without. The event-aware segmentation of U-NeXt versus the short-segment classification approach of ConvNeXt as discussed in Section \ref{sec:DevSetResults} can give some explanation for this improvement. This improvement is clinically important because seizure burden is associated with neurodevelopmental outcome \cite{kharoshankaya2016burden} and initial treatment response \cite{numis2024burden}.

The additional training data for U-NeXt+ led to substantial improvements over U-NeXt on Cork, but smaller gains on Helsinki. We suspect that this reflects Helsinki's shorter total recording duration (112 hours) and these models approaching its performance ceiling. Further improvements may be difficult to assess with the existing 3 reviewers but there might still be value in comparing to a more robust consensus formed by a larger group of reviewers.

We also tested for expert-level equivalence. There have been several published methods for such a
test that are comprehensively described and evaluated in
\cite{kljajic2026honestreliableevaluationexpert}. Here we found U-NeXt+ to be equivalent to experts on 2
variants of the most reliable test which measures the impact of replacing an expert with the AI on
the level of inter-rater agreement, $\Delta \kappa$ (see \tabref{tab:seizure-performance}). In
the first test, if $\Delta \kappa$ is statistically consistent with 0, then the model is deemed to
be expert-level i.e. failing to reject the null hypothesis that the model is equivalent (U-NeXt+
$p=0.368$ for Helsinki and $p=0.470$ for Cork; expert equivalent). A legitimate criticism is that failing to detect a difference does not establish equivalence. Alternatively, the same $\Delta \kappa$ distribution supports a non-inferiority test with a margin set by the margin of error in inter-rater agreement. In this framing, rejecting the null hypothesis of model inferiority supports expert equivalence (U-NeXt+ $p<0.001$ for both Helsinki and Cork; expert equivalent). In practical terms the former test is a constraint on the upper CI of the $\Delta \kappa$ distribution whereas the latter constrains the lower CI.

\subsection{Clinical significance of false detections}
One metric that is particularly salient to end users is the number of false detections. This is commonly reported as the hourly number of distinct prediction events with no overlap with a true event (FD/h in \tabref{tab:seizure-performance}). This is clinically relevant to alarm disruption, unnecessary specialist reviews, and treatment decision-support reliability.

However, FD/h is sensitive to small model changes and post-processing details. We can illustrate this by considering the case of a prediction that is improved to contain less false-positive time overall but fragments one false detection into several. \figref[(a)]{fig:fp_fragmentation} shows this for a real case of a false positive on Cork where the FD/h was tripled despite a reduction in total false-positive time when moving from ConvNeXt to U-NeXt. Clearly the U-NeXt prediction is preferable but selecting for FD/h only would be misleading. The focus on FD/h therefore incentivises post-processing techniques which result in long contiguous events e.g. adding a prediction "collar" \cite{temko2011performance,oshea2020fullyconvolutional,tapani2019}. This however will necessarily introduce systematic bias in seizure burden estimation. In \cite{hogan2025scaling} we argued for minimal post-processing and we further improved that here (see Sec. \ref{sec:post-processing}) because we wanted to avoid such biases and keep the models as generalisable as possible.

\PaperFigure{false-detections}

FD/h should therefore be supplemented by measures such as false-positive minutes per hour (FP min/h in \tabref{tab:seizure-performance}). More importantly, we can directly address the clinical concern of false-alarm disruption by separating alarms from detections. First, an alarm can be configured to require a minimum event duration. In addition, if staff identify a false detection, for example from artefactual EEG, they can silence subsequent alarms for a configurable period, such as 5 minutes. In \figref[(b)]{fig:false-alarms} we show some reasonable options for this configuration which can reduce alarms by approximately 50\% without compromising the underlying model through aggressive post-processing.

Inter-rater disagreement provides further context. In \figref[(c)]{fig:false-alarms}, each expert's annotation serves as the reference for the other experts and model predictions. When comparing the average over all references we see that U-NeXt+ is comparable to the experts on Cork but has approximately 50\% less inter-rater FD/h on Helsinki (albeit at the cost of lower event sensitivity). This is consistent with the expert-level equivalence test in Section \ref{sec:Results}, which suggests that even when the model disagrees with the unanimous consensus, there may be other reviewers that would agree with the model\footnote{An important qualifier here is that we have not shown equivalence in the \emph{types} of mistakes the models make compared with those made by experts. It may be the case that model failure modes lie outside the distribution of mistakes made by experts.}.

Finally, we consider the model's potential role in clinical workflows. For example, seizure burden thresholds could prompt EEG review by a neurophysiologist or support decisions about antiseizure treatment. We use 0.5 min/h, proposed for neonatal treatment trials \cite{soul2019recommendations}, and 12 min/h, used in ACNS critical-care terminology \cite{hirsch2021terminology}. We apply these thresholds to majority-consensus annotations to label 1-hour EEG segments as positive (at or above threshold) and negative (below threshold). \tabref{tab:burden-decisions} reports results for Cork and Helsinki. We see that U-NeXt is able to achieve high performance for both thresholds. Notably, on Cork, whose long-duration recordings are typical of a clinical setting, we see extremely high NPVs of 99.0\% and 99.6\% for the 0.5 min/h and 12 min/h thresholds respectively, suggesting the model could be a reliable safety net in resource constrained settings.

\begin{table}[!ht]
\centering
\caption{Seizure-burden threshold classification against majority-consensus annotations. Segments are 1~h where possible (minimum 20~min; mean 59.4~min Cork, 53.9~min Helsinki). Error MAE: mean absolute predicted--reference burden difference (min/h), over FP/FN segments only.}
\label{tab:burden-decisions}
\setlength{\tabcolsep}{0pt}
\setlength{\arrayrulewidth}{0.3pt}
\renewcommand{\arraystretch}{1.1}
\settowidth{\dimen0}{Helsinki SVM}\edef\burdenmodelwidth{\the\dimen0}
\settowidth{\dimen0}{Error MAE}\edef\burdenmaewidth{\the\dimen0}
\settowidth{\dimen0}{$\geq0.5$\,min/h}\edef\burdensidewidth{\the\dimen0}
\dimen0=\dimexpr(\linewidth-\burdenmodelwidth-\burdenmaewidth-\burdensidewidth-28pt-2\arrayrulewidth)/9\relax
\edef\burdennumwidth{\the\dimen0}
\newcommand{\burdennum}[1]{\makebox[\burdennumwidth][c]{#1}}
\makeatletter
\newcommand{\burdengrouptop}{\rule{0pt}{\dimexpr\ht\@arstrutbox+1.5pt\relax}}
\newcommand{\burdengroupbottom}{\rule[-\dimexpr\dp\@arstrutbox+1.5pt\relax]{0pt}{\dimexpr\ht\@arstrutbox+\dp\@arstrutbox+1.5pt\relax}}
\makeatother
\begin{tabular}{@{}l@{\hspace{6pt}}l@{\hspace{6pt}}*{4}{c}@{\hspace{4pt}}|@{\hspace{4pt}}*{5}{c}@{\hspace{4pt}}|@{\hspace{4pt}}c@{}}
\noalign{\hrule height 0.8pt}
\makebox[\burdensidewidth]{} & & \burdennum{TP} & \burdennum{FP} & \burdennum{FN} & \burdennum{TN} & \burdennum{Sens.} & \burdennum{Spec.} & \burdennum{PPV} & \burdennum{NPV} & \burdennum{MCC} & \begin{tabular}[c]{@{}c@{}}Error MAE\\(min/h)\end{tabular} \\
\noalign{\hrule height 0.6pt}
\multirow{3}{*}{\shortstack[l]{\textbf{Cork}\\$\geq0.5$\,min/h}}\burdengrouptop & ConvNeXt & \burdennum{301} & \burdennum{129} & \burdennum{29} & \burdennum{2103} & \burdennum{91.2} & \burdennum{94.2} & \burdennum{70.0} & \burdennum{98.6} & \burdennum{0.766} & 6.53 \\
 & U-NeXt & \burdennum{304} & \burdennum{140} & \burdennum{26} & \burdennum{2092} & \burdennum{92.1} & \burdennum{93.7} & \burdennum{68.5} & \burdennum{98.8} & \burdennum{0.760} & 4.84 \\
\burdengroupbottom & U-NeXt+ & \burdennum{309} & \burdennum{96} & \burdennum{21} & \burdennum{2136} & \burdennum{\textbf{93.6}} & \burdennum{\textbf{95.7}} & \burdennum{\textbf{76.3}} & \burdennum{\textbf{99.0}} & \burdennum{\textbf{0.820}} & \textbf{3.18} \\
\noalign{\hrule height 0.25pt}
\multirow{3}{*}{\shortstack[l]{\textbf{Cork}\\$\geq12$\,min/h}}\burdengrouptop & ConvNeXt & \burdennum{61} & \burdennum{37} & \burdennum{14} & \burdennum{2450} & \burdennum{81.3} & \burdennum{98.5} & \burdennum{62.2} & \burdennum{99.4} & \burdennum{0.702} & 21.06 \\
 & U-NeXt & \burdennum{63} & \burdennum{36} & \burdennum{12} & \burdennum{2451} & \burdennum{84.0} & \burdennum{98.6} & \burdennum{63.6} & \burdennum{99.5} & \burdennum{0.722} & 17.22 \\
\burdengroupbottom & U-NeXt+ & \burdennum{65} & \burdennum{20} & \burdennum{10} & \burdennum{2467} & \burdennum{\textbf{86.7}} & \burdennum{\textbf{99.2}} & \burdennum{\textbf{76.5}} & \burdennum{\textbf{99.6}} & \burdennum{\textbf{0.808}} & \textbf{10.59} \\
\noalign{\hrule height 0.7pt}
\multirow{4}{*}{\shortstack[l]{\textbf{Helsinki}\\$\geq0.5$\,min/h}}\burdengrouptop & Helsinki SVM & \burdennum{56} & \burdennum{21} & \burdennum{6} & \burdennum{35} & \burdennum{\textbf{90.3}} & \burdennum{62.5} & \burdennum{72.7} & \burdennum{85.4} & \burdennum{0.554} & 6.54 \\
 & ConvNeXt & \burdennum{50} & \burdennum{5} & \burdennum{12} & \burdennum{51} & \burdennum{80.6} & \burdennum{\textbf{91.1}} & \burdennum{90.9} & \burdennum{81.0} & \burdennum{0.718} & 10.58 \\
 & U-NeXt & \burdennum{53} & \burdennum{5} & \burdennum{9} & \burdennum{51} & \burdennum{85.5} & \burdennum{\textbf{91.1}} & \burdennum{91.4} & \burdennum{85.0} & \burdennum{0.765} & 5.08 \\
\burdengroupbottom & U-NeXt+ & \burdennum{55} & \burdennum{5} & \burdennum{7} & \burdennum{51} & \burdennum{88.7} & \burdennum{\textbf{91.1}} & \burdennum{\textbf{91.7}} & \burdennum{\textbf{87.9}} & \burdennum{\textbf{0.797}} & \textbf{3.17} \\
\noalign{\hrule height 0.25pt}
\multirow{4}{*}{\shortstack[l]{\textbf{Helsinki}\\$\geq12$\,min/h}}\burdengrouptop & Helsinki SVM & \burdennum{14} & \burdennum{3} & \burdennum{16} & \burdennum{85} & \burdennum{46.7} & \burdennum{96.6} & \burdennum{82.4} & \burdennum{84.2} & \burdennum{0.536} & 17.33 \\
 & ConvNeXt & \burdennum{18} & \burdennum{2} & \burdennum{12} & \burdennum{86} & \burdennum{60.0} & \burdennum{97.7} & \burdennum{90.0} & \burdennum{87.8} & \burdennum{0.670} & 14.49 \\
 & U-NeXt & \burdennum{19} & \burdennum{1} & \burdennum{11} & \burdennum{87} & \burdennum{63.3} & \burdennum{\textbf{98.9}} & \burdennum{95.0} & \burdennum{88.8} & \burdennum{0.722} & 8.99 \\
\burdengroupbottom & U-NeXt+ & \burdennum{23} & \burdennum{1} & \burdennum{7} & \burdennum{87} & \burdennum{\textbf{76.7}} & \burdennum{\textbf{98.9}} & \burdennum{\textbf{95.8}} & \burdennum{\textbf{92.6}} & \burdennum{\textbf{0.817}} & \textbf{6.85} \\
\noalign{\hrule height 0.8pt}
\end{tabular}
\end{table}

\UseHook{paper/layout/before-inference}
\subsection{Inference performance}\label{sec:inference-performance}
We compared model inference costs to quantify U-NeXt's efficiency gains. FLOP counts (\tabref{tab:model_variants}) allow hardware-independent comparison but do not convey practical speedups or account for ConvNeXt's $>$98\% inference overlap, which U-NeXt does not require. To make it more tangible we benchmarked seizure-model batch inference on 1 hour of the 18-channel Helsinki EEG. We benchmarked PyTorch Lightning implementations of ConvNeXt and U-NeXt on the CPU and GPU of a 2023 MacBook Pro M2 Max. The resulting $\sim200\times$ gains are summarised in \figref[(a)]{fig:segment_size}. For 1 hour of 18-channel EEG the total compute was 1.14 TFLOPs for U-NeXt vs 268.1 for ConvNeXt. These totals use 259,200 overlapping 16~s windows for ConvNeXt (0.25~s stride) and, per channel, fourteen 256~s windows plus a final 16~s window for U-NeXt, giving a 236-fold compute reduction. Median inference times over 3 runs were 30~s vs 2~h 7~min on CPU or 0.78~s vs 2~min 37~s on GPU. For comparison the open-source MATLAB implementation of the Helsinki SVM took 4.5 hours on CPU. Although this is likely not optimised, due to the nature of the feature calculation it is not readily amenable to GPU acceleration, which is trivial for CNNs with modern frameworks. As a result, we did not evaluate the Helsinki SVM on Cork because it would have taken $\sim212$ days to complete.

Both ConvNeXt and U-NeXt can process multi-channel EEG faster than it is recorded, but ConvNeXt may require a GPU or multiple CPUs processing channels in parallel. U-NeXt achieves real-time inference on one CPU by producing 4~Hz predictions in a single forward pass that would require many forward passes for ConvNeXt.

These comparisons were made for batch processing. Live inference would benefit from more frequent updates, with some lag to limit pre-processing filter edge effects and allow smoothing. Classification models such as ConvNeXt also require lag because predictions nominally refer to segment centres: a 16~s window requires 8~s of future EEG.

U-NeXt's fully convolutional architecture allows variable segment sizes at inference. \figref[(b)]{fig:segment_size} shows that longer segments give modest performance improvements, likely because they provide more temporal context, further supporting the move away from short-segment classification. A deployed model should therefore retain some recent EEG context, which will reintroduce some overlap and increase computational cost.

\PaperFigure{inference}

\UseHook{paper/layout/before-reduced-montage}
\subsection{Reduced montage detection}
Helsinki and Cork use 18- and 8-channel montages, respectively, whereas long-duration neonatal clinical monitoring commonly uses reduced channel montages such as 2 channels for amplitude-integrated EEG (aEEG) systems. As was the case for ConvNeXt, the per-channel training lets U-NeXt accommodate different montages. In \figref{fig:reduced-montage} we show the performance of U-NeXt+ on Cork with different reduced channel configurations. As expected, we see some degradation in sensitivity due to localised seizures occurring outside the retained channels, but even with just two central channels, the model retains more than 80\% of the eight-channel event sensitivity. Notably, overall MCC (\figref[(c)]{fig:reduced-montage}) remains remarkably stable because the increase in PPV offsets the decrease in sensitivity. This analysis suggests that the model can be usefully deployed in reduced channel settings with the important caveat that focal seizures outside the retained channels can be missed.

\PaperFigure{reduced-montage}

\section{Discussion}
We have introduced U-NeXt, a new general-purpose event detector for neonatal EEG. It improves upon our previous ConvNeXt models, which are state of the art in both LVI \cite{otoole2026expert} and seizure detection \cite{hogan2025scaling}, along several axes. First, it comprehensively outperforms the ConvNeXt model on 3 different tasks: LVI, artefact, and seizure detection. It does this while also being far more general, without the need to tune the architecture, hyperparameters, or post-processing to a particular task. It sets a new state-of-the-art result for neonatal seizure detection on Helsinki, achieves expert-level performance on both Helsinki and Cork, and produces high-fidelity estimates of key clinical markers like seizure burden \cite{kharoshankaya2016burden,numis2024burden}. Finally, we demonstrate a significantly more computationally efficient model with $\sim$250$\times$ speedup vs ConvNeXt models for batch inference.

Much of the literature has focused on treating event detection as a classification problem
through short-time analysis. By contrast, experts perform semantic segmentation by annotating event start and end times in continuous EEG. Rather than discard this detail, we use these sample annotations directly as a rich training signal. The resulting model detection masks are much more human-like than those produced by previous approaches with smooth, coherent detections and accurate start and end times. This also improved detection sensitivity across event types, including short events that are challenging for experts and automated systems alike \cite{stevenson2015interobserver}.

As in our previous work, we use per-channel modelling, which, although costly, improves training-signal quality and prediction flexibility. We demonstrate that with just 2 of 8 channels, U-NeXt retained remarkably stable MCC and more than 80\% of the eight-channel event sensitivity, although focal seizures outside the retained channels could be missed.

We also explored the practical clinical implications of reported false detections. Although important and of great interest to clinicians, FD/h can misrepresent performance and clinical impact. False detections should not be equated with false alarms: FD/h can overestimate the false-alarm rate by a factor of 2 or more. We also highlighted that it should always be reported in the context of inter-rater false detections, which we found to be comparable to U-NeXt on Cork but twice as large on Helsinki. Finally, when used in the context of decision support based on seizure burden, U-NeXt had remarkably good performance with a sensitivity/specificity of 86.7\%/99.2\% at a threshold of 12~min/h on Cork. This suggests that U-NeXt could be a valuable adjunct decision support tool for cotside caregivers when EEG expertise is unavailable or delayed.

This study is not without limitations. Despite its improved generality, the framework still requires a separate model for each task. Disjoint annotations and differences in dataset scale complicate joint training. Training one large model, for example for seizure detection, then fine-tuning for other tasks may be feasible, but was not explored here. Another area for improvement would be testing with additional event types to further assess the framework's generality. We selected LVI, artefact, and seizure detection for their clinical utility and diverse durations and morphologies, although many event types remain untested. The artefact dataset itself has events that vary widely in duration and morphology, ranging from brief transients to sustained periodic patterns and irregular, noise-like activity. It is promising, therefore, that this architecture would also work well on other event types.

These results demonstrate the value of end-to-end learning, generality, and scale. Removing hand-crafted features, overly customised post-processing, and task-specific model adaptations can produce a high-performance, general-purpose event detector design. This can then adapt quickly to new tasks without architectural redesign.

  \section*{Acknowledgements}
  \PaperFundingStatement\space\PaperFunderRole

  \section*{Competing Interests}
  \PaperCompetingInterests

\end{document}